\documentclass[a4paper, 10pt, conference]{IEEEtran}
\IEEEoverridecommandlockouts
\usepackage{cite}
\usepackage{amsmath,amssymb,amsfonts}
\usepackage{algorithmic}
\usepackage[dvipdfmx]{graphicx}
\usepackage{textcomp}
\usepackage{xcolor}
\usepackage{bm}
\usepackage{subcaption}
\usepackage{algorithm}
\usepackage{comment}
\usepackage{color}

\DeclareMathOperator*{\argmin}{\arg\!\min}

\newcommand{\vect}[1]{\boldsymbol{#1}}

\def\BibTeX{{\rm B\kern-.05em{\sc i\kern-.025em b}\kern-.08em
    T\kern-.1667em\lower.7ex\hbox{E}\kern-.125emX}}
\begin{document}

\title{Adversarial Body Shape Search \\for Legged Robots\\
\thanks{This work was supported by JSPS KAKENHI Grant Numbers JP19K12039 and JP22H03658.}
}

\author{\IEEEauthorblockN{Takaaki Azakami}
\IEEEauthorblockA{
Graduate School of Science and Engineering \\
Chiba University \\
Chiba, Japan \\
Email: takaaki.azakami@chiba-u.jp}

\and
\IEEEauthorblockN{Hiroshi Kera}
\IEEEauthorblockA{
Graduate School of Engineering \\
Chiba University \\
Chiba, Japan \\
Email: kera@chiba-u.jp
}

\and
\IEEEauthorblockN{Kazuhiko Kawamoto}
\IEEEauthorblockA{
Graduate School of Engineering \\
Chiba University \\
Chiba, Japan \\
Email: kawa@faculty.chiba-u.jp
}

}

\maketitle

\begin{abstract}
We propose an evolutionary computation method for an adversarial attack
on the length and thickness of parts of legged robots by deep reinforcement learning.
This attack changes the robot body shape and interferes with walking—we call the attacked
body as \textit{adversarial body shape}.
The evolutionary computation method searches adversarial body shape by minimizing
the expected cumulative reward earned through walking simulation.
To evaluate the effectiveness of the proposed method, 
we perform experiments with three legged robots, Walker2d, Ant-v2, and Humanoid-v2 in OpenAI Gym.
The experimental results reveal that Walker2d and Ant-v2 are more vulnerable to the attack
on the length than the thickness of the body parts, whereas Humanoid-v2 is vulnerable to the attack on both of the length and thickness. 
We further identify that the adversarial body shapes break left-right symmetry or 
shift the center of gravity of the legged robots.
Finding adversarial body shape can be used to proactively diagnose the vulnerability of
legged robot walking.
\end{abstract}

\begin{IEEEkeywords}
Adversarial attack, Deep reinforcement learning, Legged robot control
\end{IEEEkeywords}

\section{Introduction}
Deep reinforcement learning in robotics has been widely studied \cite{10.1007/s11370-021-00398-z}; the vulnerability and robustness have also been attracting attention. In particular, the vulnerability to adversarial attacks is inherent to deep reinforcement learning, and improving the robustness is a considerable challenge
\cite{xiao2019characterizing,9536399}.
Among the adversarial attacks in robotics, the attack on state observations is a main topic, because they directly affect the control of robots.

\begin{figure}[t]
    \begin{center}
        \includegraphics[width=\linewidth]{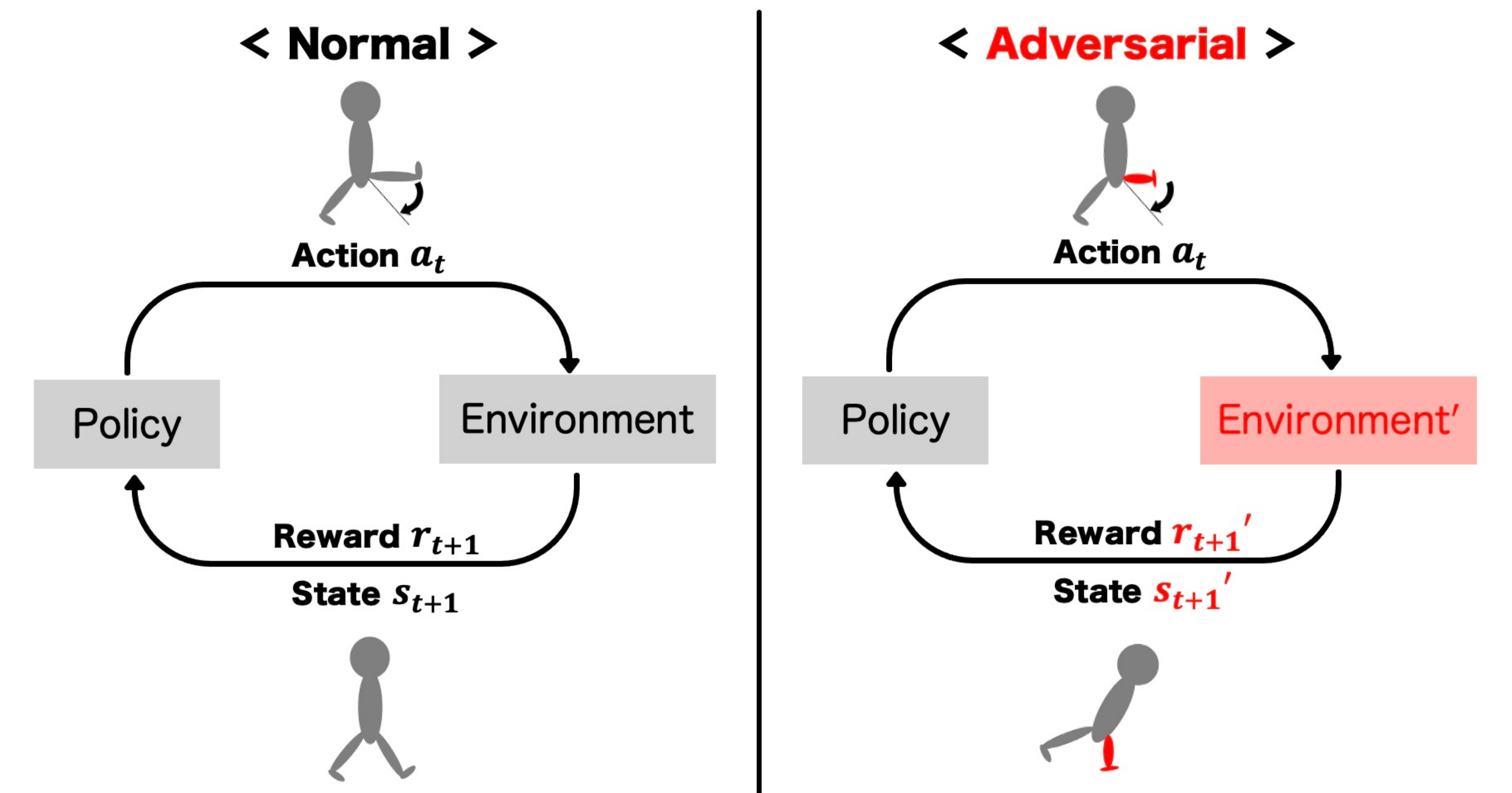}
    	\caption{Adversarial body shape search}
    	\label{overall}
    \end{center}
\end{figure}

In this study, we consider the adversarial attack on the body shape of legged robots, which is seen as an attack on environments, not the state observation. 
The attack adds adversarial perturbations to the length and thickness of the legged robots;
we call the attacked body shape as \textit{adversarial body shape}.
Figure \ref{overall} illustrates the adversarial body shape search, where the adversarial attack
makes the length of the bipedal robot’s leg shorter and throws the robot off balance. 
Robot body shape changes can occur owing to various factors such as oxidation of the metal material, adhesion of foreign matter to the surface, and formation of defects and dents caused by collision. If such changes are too small to detect, vulnerability is a potential risk. The adversarial attack can search small perturbations that causes walking instability of the legged robots.
Thus, the adversarial body shape search can be used
to proactively diagnose the vulnerability of robot walking.

\begin{figure}[t]
  \begin{center}
    \begin{tabular}{c}
      \begin{minipage}[h]{0.33\columnwidth}
    		\centering
    		\includegraphics[width=0.95\columnwidth]{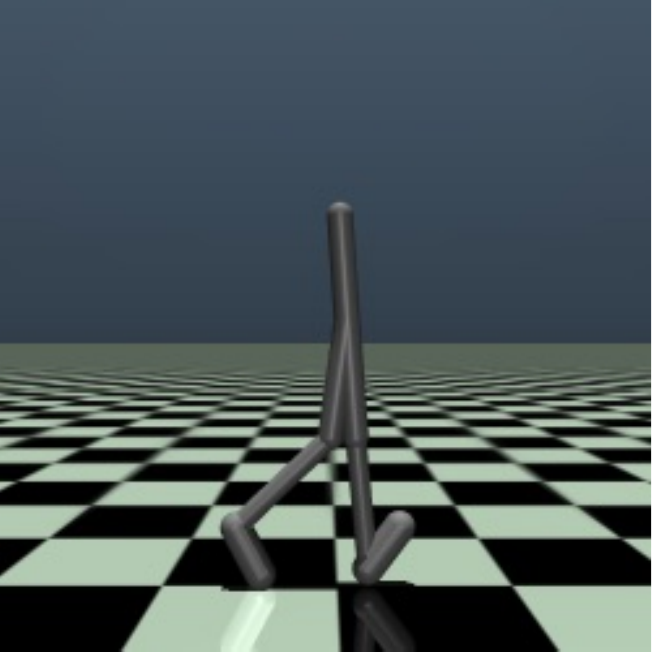}
    		\label{walker2d_length_non_eps}\\
        \subcaption{Walker2d-v2}
    	\end{minipage}%
    	\begin{minipage}[h]{0.44\columnwidth}
    		\centering
    		\includegraphics[width=0.95\columnwidth]{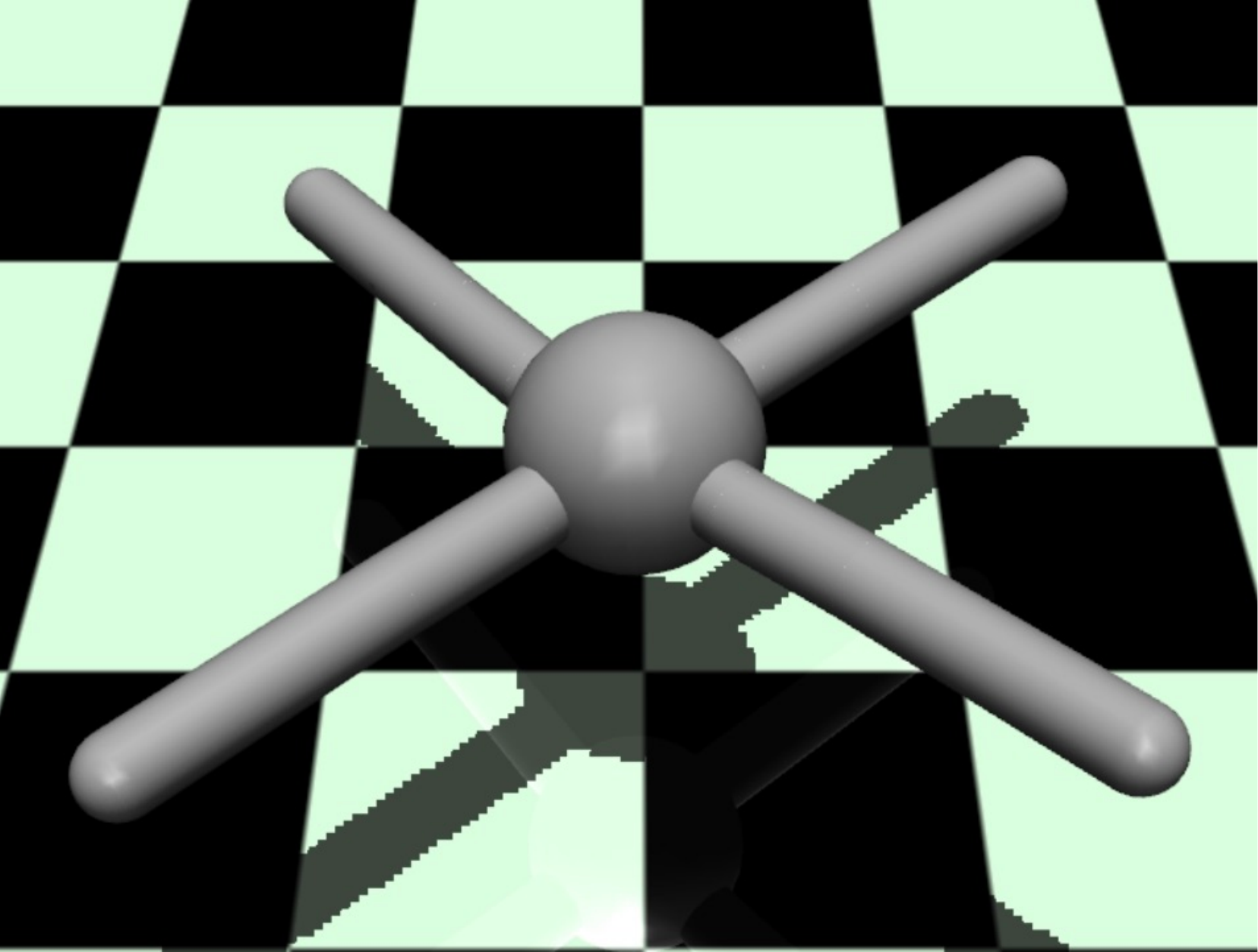}
    		\label{ant_length_non_eps}\\
        \subcaption{Ant-v2}
    	\end{minipage}%
    	\begin{minipage}[h]{0.26\columnwidth}
    		\centering
    		\includegraphics[width=0.71\columnwidth]{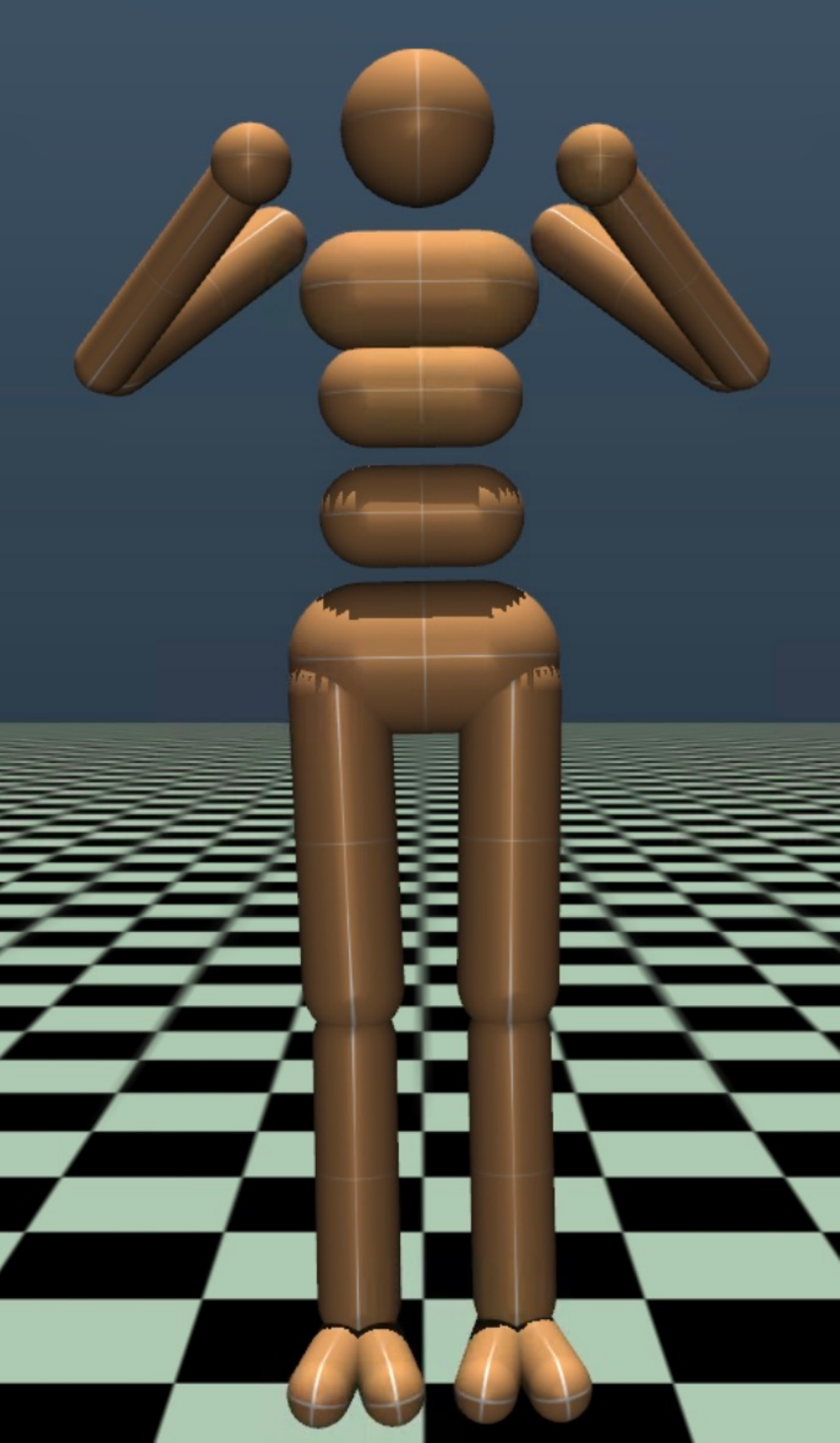}
    		\label{humanoid_length_non_eps}\\
        \subcaption{Humanoid-v2}
    	\end{minipage}%
    \end{tabular}
    \caption{Legged robots}
    \label{robot}
  \end{center}
\end{figure}

We propose a differential evolution method \cite{storn1997differential} for searching adversarial body shapes of the legged robots. The proposed search method is designed to reduce the cumulative reward earned by walking simulation of the legged robots. The rewards can be obtained only through walking simulation. Therefore, we employ evolutionary computation with walking simulation on a physical engine to search adversarial body shapes. 
We perform experiments with three legged robots: the bipedal robot Walker2d-v2, the quadruped robot Ant-v2, and the bipedal robot Humanoid-v2 \cite{brockman2016openai}, as shown in Fig.~\ref{robot}. 
These robot environments run on the physical engine MuJoCo \cite{todorov2012mujoco}.
We evaluate the effectiveness of the proposed search
method in terms of the average cumulative rewards for
1000 walking simulations.
The experimental results reveal that Walker2d and Ant-v2
are more vulnerable to the attack on the length
than the thickness of the body parts, whereas, Humanoid-v2
is vulnerable to the attack on both of the length and thickness.
In addition, we investigate the adversarial perturbations that interfere with 
the walking task for each body part and discover that 
the perturbations break left-right symmetry or shift the center of gravity of the legged robots.

The main contributions of this study are as follows:
\begin{itemize}
 \item We propose an evolutionary computation method with physical walking simulations
 for searching adversarial body shapes of legged robots. This method will be used to proactively diagnose the vulnerability of the legged robots.
 \item We discover the adversarial body shapes that interfere with the walking task for the first time.
\end{itemize}

\section{Related work}
Robot design optimization for legged robots has been studied.
Several studies \cite{Ha2019ReinforcementLF,Schaff2019JointlyLT,pmlr-v100-luck20a}
jointly optimize robot design and reinforcement learning.
Using REINFORCEMENT,
Ha et al.~\cite{Ha2019ReinforcementLF} improve policies in environments such as BipedalWalker-2d and Ant-v1, and design more task-appropriate bodies.
Schaff et al.~\cite{Schaff2019JointlyLT} retain the distribution on the length and thickness of
the legged robots: Hopper, Walker2d, and Ant. 
They use reinforcement learning to optimize the control policy by
maximizing the expected reward for the design distribution.
Luck et al.~\cite{pmlr-v100-luck20a} combine design optimization and reinforcement learning to minimize the number of prototypes for optimal robot foot design, improving data efficiency.
These studies optimize the robot design by maximizing the rewards.
However, they do not consider the vulnerability inherent to the legged robots trained by
reinforcement learning.
Wang et al.~\cite{wang2018neural} use evolutionary computation
to design an optimal size and position of fish fins.
To design that, neural graph evolution \cite{wang2018nervenet} is used, and the method
iteratively evolves the graph structure using mutations.
Desai et al.~\cite{DBLP:journals/corr/abs-1801-00385} propose 
an interactive computational design system that enables users to
design legged robots with desired morphologies and behaviors.
The interactive system automatically suggests candidate robot designs 
to achieve a specified behavior or task performance by an optimization algorithm.
Although these studies do not use reinforcement learning, 
they still do not consider the vulnerability of the legged robots.
In this study, we propose a method for detecting the vulnerability inherent to 
the body shapes of the legged robots trained by deep reinforcement learning.

\section{Adversarial body shape search}

This section describes the adversarial body shape perturbation and the adversarial body shape generation algorithm.

\begin{algorithm}[t]
    \caption{Adversarial Body Shape Generation Algorithm}
    \label{alg1}
    \begin{algorithmic}[1]
    \REQUIRE Population size $NP$
    \STATE Generate initial population of body shape perturbations $\{\vect{\delta}_0^{(1)},\ldots,\vect{\delta}_0^{(NP)}\}$ with $\delta_{j,0}^{(i)}\sim U\left([-\epsilon,\epsilon]\right)$ 
    \STATE $G_{\mathrm{min}} \leftarrow \infty$
    \FOR {each generation $g = 1,2,\ldots$}
	\FOR {each individual $i = 1,2,\ldots,NP$}
		\STATE Compute adversarial shape $\bm{b}^{(i)}_\mathrm{adv}=\bm{b}+\bm{\delta}_{g-1}^{(i)}\odot\bm{b}$
	    \STATE Generate trial individual $\bm{u}_{g}^{(i)}$ by mutation and crossover in Eqs.~(\ref{equ: mutation}) and (\ref{equ: crossover})
	    \STATE Compute trial shape $\bm{b}^{(i)}_\mathrm{trial}=\bm{b}+\bm{u}_{g}^{(i)}\odot\bm{b}$
    	\STATE Compute average cumulative rewards in Eq.~(\ref{equ: average cumulative reward}) for\\
    	trial $\bar{G}(\bm{b}^{(i)}_\mathrm{trial})$
        and target $\bar{G}(\bm{b}^{(i)}_\mathrm{adv})$
        \IF {$\bar{G}(\bm{b}^{(i)}_\mathrm{trial}) \leq \bar{G}(\bm{b}^{(i)}_\mathrm{adv})$}
        \STATE Accept trial individual $\vect{\delta}_{g}^{(i)} \leftarrow \vect{u}_{g}^{(i)}$
        \IF {$\bar{G}(\bm{b}^{(i)}_\mathrm{trial})<G_{\mathrm{min}}$}
        \STATE $G_{\mathrm{min}}\leftarrow \bar{G}(\bm{b}^{(i)}_\mathrm{trial})$
        \STATE Update best individual $\vect{\delta}_{\mathrm{best}}\leftarrow \vect{\delta}_g^{(i)}$
        \ENDIF
	\ELSE
        \STATE Retain target individual $\vect{\delta}_{g}^{(i)} \leftarrow \vect{\delta}_{g-1}^{(i)}$ 
	\ENDIF
	\ENDFOR
    \ENDFOR
    \RETURN $\bm{\delta}_{\mathrm{best}}$
    \end{algorithmic}
\end{algorithm}
\subsection{Adversarial Body Shape Perturbation}\label{ABSP}

We denote $\bm{b}$ as a body shape vector consisting of the length or thickness of 
the body parts of legged robots in Fig.~\ref{robot}.
We assume that body shape $\bm{b}$ is constant over time.
Body shape $\bm{b}$ affects reward ${r}_t$
through reward function $r(\cdot)$, i.e.,
\begin{align}
{r}_t=r(\bm{s}_t,\bm{a}_t,\bm{b})
\end{align}
where, 
$\bm{a}_t$ is the action vector of a legged robot at time $t$,
and reward $r_t$ is computed by running a physical simulation for robot walking. 

In this study, we perturb body shape $\bm{b}$ as
\begin{align}
\bm{b}_{\mathrm{adv}}=\bm{b}+\bm{\delta}\odot\bm{b}
\label{equ: body shape perturbation}
\end{align}
where $\bm{\delta}$ is an adversarial perturbation of body shape $\bm{b}$,
and $\odot$ is the Hadamard product.
Equation (\ref{equ: body shape perturbation}) perturbs the ratio of the length or thickness
of the parts of legged robots. 
For example, the length of the $i$-th body part, denoted by $b_i$, is perturbed by
\begin{equation}
    b_{i,\mathrm{adv}} = b_i + \delta_i b_i = (1+\delta_i)b_i
\end{equation}
Adversarial perturbation $\bm{\delta}$
is computed by minimizing the expected cumulative reward as
\begin{align}
\bm{\delta}_\mathrm{best} 
= \argmin_{\bm{\delta}} E[{G}(\bm{b}_\mathrm{adv})],\  \text{subject to}
||\bm{\delta}||_{\infty} \leq \epsilon
\label{equ: expected cumulative reward}
\end{align}
where $||\cdot||_\infty$ indicates the maximum norm, $\epsilon$ is a small positive constant, 
and $G(\bm{b}_{\mathrm{adv}})$ is the cumulative reward attacked by adversarial
perturbation $\bm{\delta}$ and is defined as
\begin{align}
    G(\bm{b}_{\mathrm{adv}}) =\sum_{t=0}^{T-1}r(\bm{s}_t,\bm{a}_t,\bm{b}_{\mathrm{adv}}).
\end{align}
The expectation in Eq.~(\ref{equ: expected cumulative reward}) is
taken with respect to a trajectory of states and actions
$\{s_0, a_0, s_1, a_1, \ldots, s_{T-1}, a_{T-1}\}$,
and 
the expected cumulative reward can be empirically estimated by running $M$ waking simulations as
\begin{equation}
E[G(\bm{b}_{\mathrm{adv}})]\approx
\bar{G}(\bm{b}_\mathrm{adv})=\frac{1}{M}\sum_{m=1}^M
G^{(m)}(\bm{b}_\mathrm{adv}),
\label{equ: average cumulative reward}
\end{equation}
where $G^{(m)}(\bm{b}_{\mathrm{adv}})$ is the reward obtained by the $m$-th walking simulation.
In experiments, we set $M=50$.

\begin{figure*}[t]
  \begin{center}
    \begin{tabular}{c}
    	\begin{minipage}[h]{0.145\linewidth}
    		\centering
    		\includegraphics[width=0.98\columnwidth]{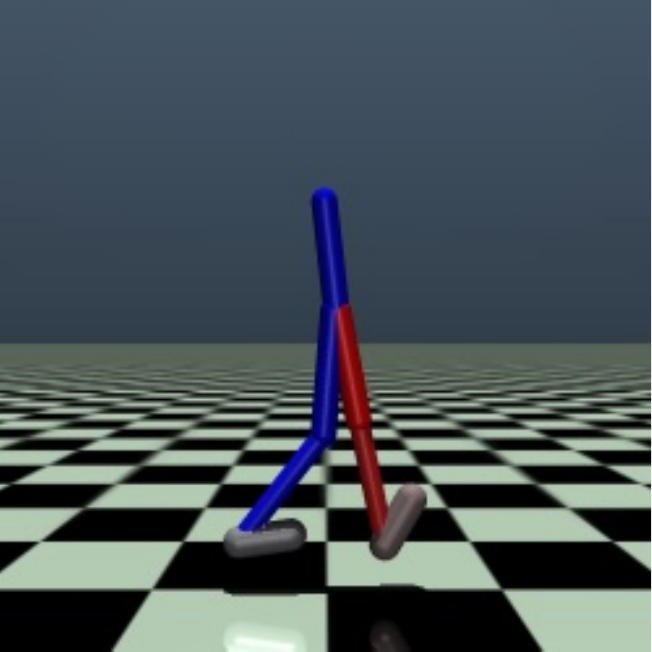}
        \subcaption{length}
    		\includegraphics[width=0.98\columnwidth]{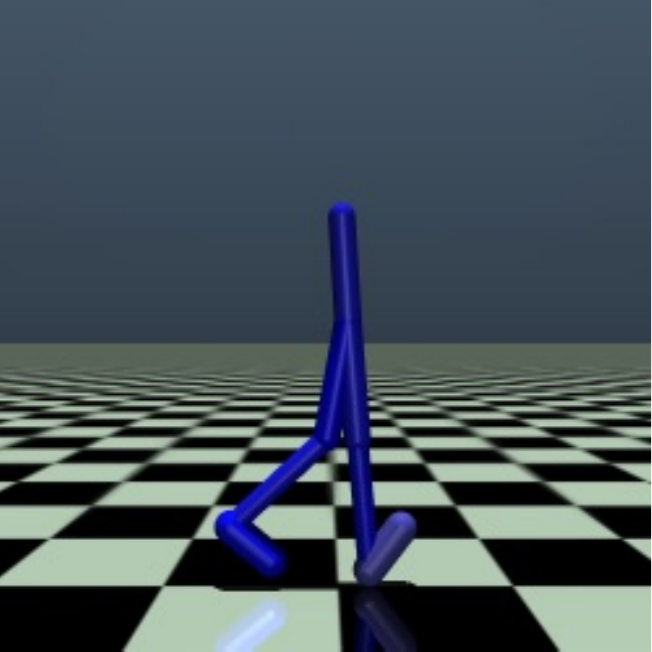}
        \subcaption{thickness}
    	\end{minipage}%
    	\begin{minipage}[h]{0.855\linewidth}
    		\centering
    	    \includegraphics[width=0.98\columnwidth]{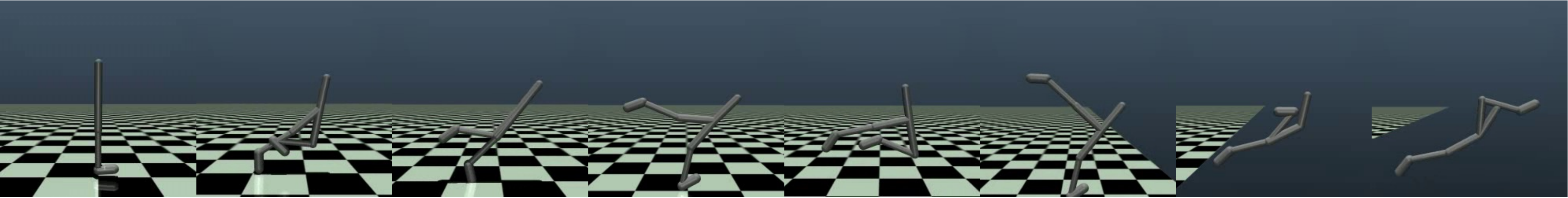}
            \includegraphics[width=0.98\columnwidth]{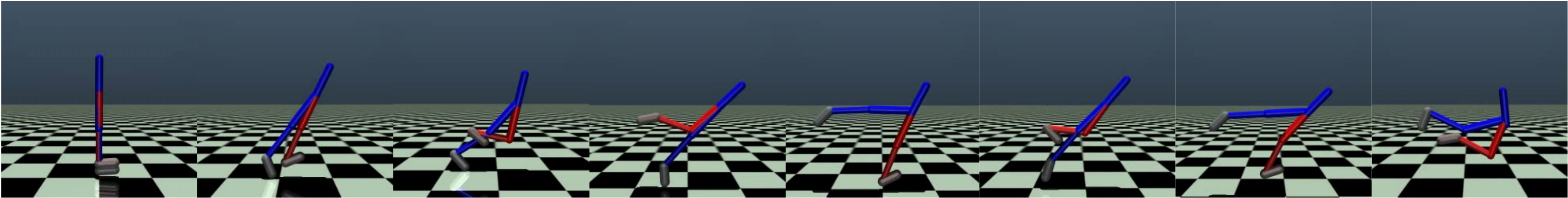}
            \includegraphics[width=0.98\columnwidth]{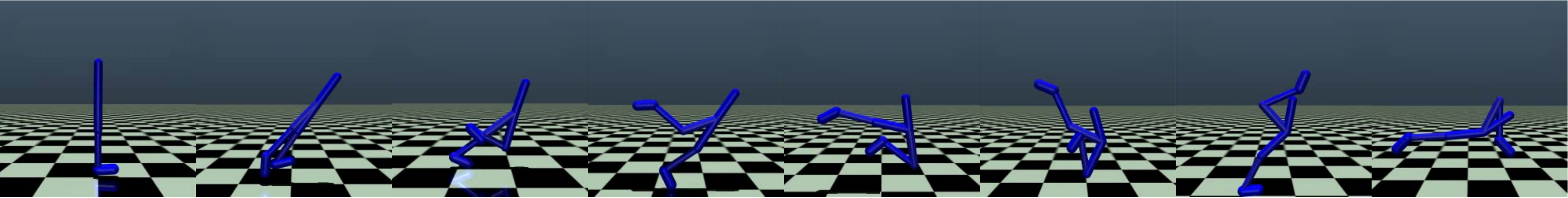}
       \subcaption{Walking animations: clean  (top), length attack  (middle), thickness attack  (bottom)}
       \label{multi pic walker2d}
       \end{minipage}
    \end{tabular}
    \caption{Adversarial body shapes with length and thickness perturbations for Walker2d-v2. }
    \label{walker2d}
  \end{center}
\end{figure*}

\begin{figure}[t]
    \begin{center}
        \includegraphics[width=0.9\linewidth]{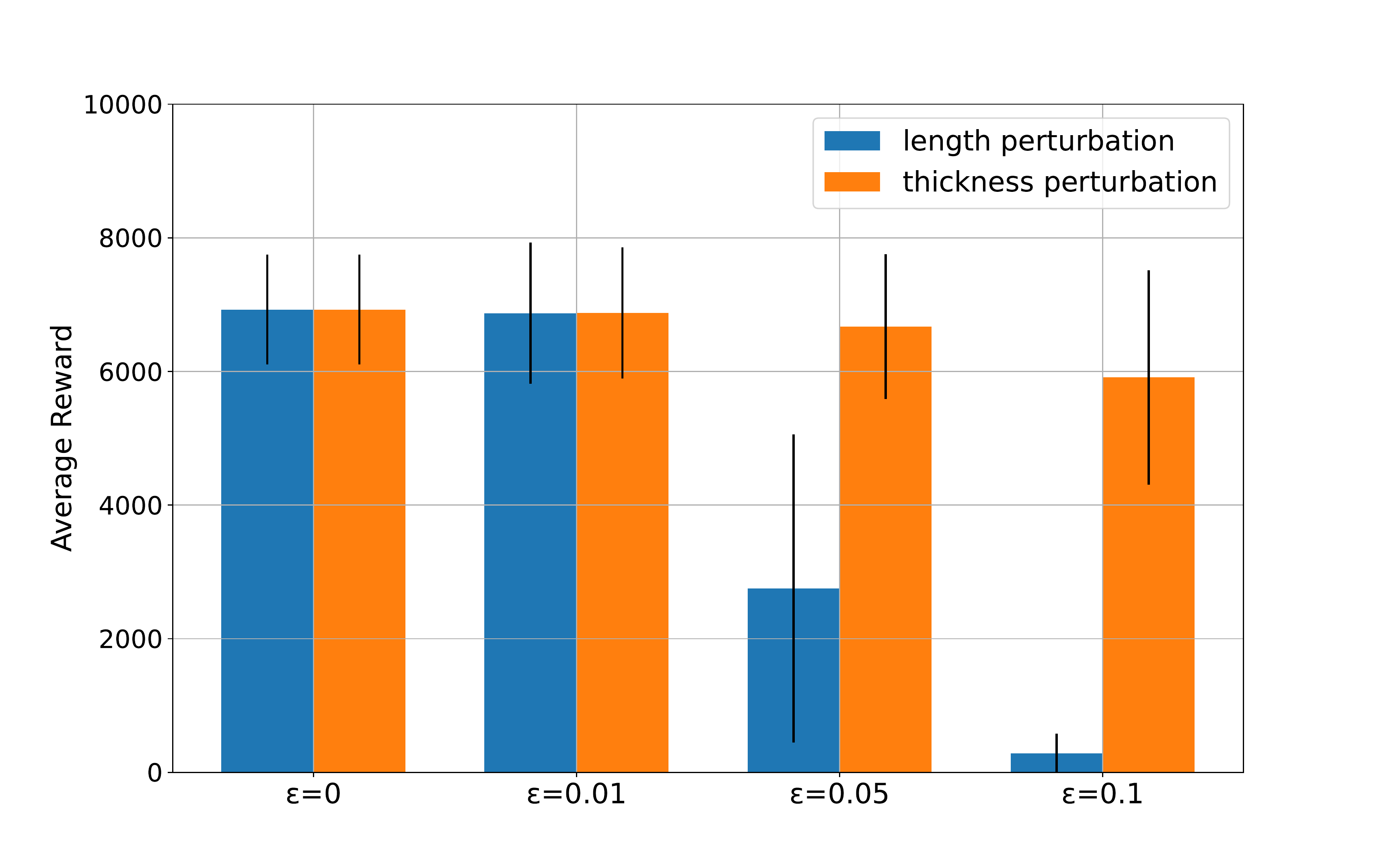}
    	\caption{Averages of the average cumulative rewards
    	with attack strength $\epsilon=0.0,0.0.1,0.05,0.1$
    	for Walker2d-2d}
    	\label{bar walker2d}
    \end{center}
\end{figure}

\subsection{Evolutionary computation for Adversarial Body Shape}\label{AA}
We employ a differential evolution method \cite{storn1997differential} to find the best perturbation 
$\bm{\delta}_\mathrm{best}$ in Eq.~(\ref{equ: expected cumulative reward}).
Note that it is difficult to compute the gradient of the cumulative reward, $G(\bm{b})$,
with respect to $\bm{b}$ because the computation of $G(\bm{b})$ requires physical simulations.
Therefore, we cannot use efficient methods such as gradient descent.
Our simulation-based adversarial attack is a black box attack \cite{9536399}.

We show the differential evolution method for adversarial body shape attack
in \textbf{Algorithm} \ref{alg1}.
For differential evolution, we first generate an initial population of body shape perturbations
$\{\bm{\delta}_0^{(1)},\ldots,\bm{\delta}_0^{(NP)}\}$,
where $NP$ is the population size.
Each element of initial individual $\bm{\delta}_0^{(i)}$ is generated according to uniform distribution $U([-\epsilon,\epsilon])$.
In differential evolution, the mutation and crossover of the population are performed.
Mutant individual $\bm{v}_{g+1}^{(i)}$ at the $g+1$-th generation 
is generated according to
\begin{equation}
    \bm{v}_{g+1}^{(i)} = \bm{\delta}_{\mathrm{best}} + F(\bm{\delta}^{(r_1)}_g-\bm{\delta}^{(r_2)}_g),
    \label{equ: mutation}
\end{equation}
where $F$ is the parameter that adjusts the rate of mutation, and $r_1,r_2$ are randomly selected
from the individual index set, $\{1,\ldots,NP\}$.
The best individual $\vect{\delta}_{\mathrm{best}}$ in Eq.~(\ref{equ: mutation})
is the intermediate best until the $g$-th generation.
The trial individual by crossover $\bm{u}_{g+1}^{(i)}$ is generated according to
\begin{equation}
{u}_{j,g+1}^{(i)} = \left\{
\begin{array}{ll}
{v}_{j,g+1}^{(i)} & \text{if}\quad r \leq CR\ \text{or}\ j=j_{r}\\
{\delta}_{j,g}^{(i)} & \text{otherwise}
\end{array}
\right.
\label{equ: crossover}
\end{equation}
where subscript $j$ indicates the $j$-the element of the vector, 
$r$ is a uniform random number in $[0,1]$,
$CR=0.7$ is a crossover constant,
and $j_r$ is a random element index.

\begin{table}[t]
    \caption{Adversarial body shape perturbation for thickness and length of Walker2d-v2}
    \label{ratio walker2d}
    \centering
    \begin{tabular}{|l||r|r|r|} \hline
    		body part	& length (\%)	& thickness (\%)	\\ \hline \hline
    		torso 			& \color{blue}+4.73 	& \color{blue}+3.93 	\\ \hline
    		right thigh 			& \color{blue}+4.82 & \color{blue}+4.91\\ 
    		right leg			 & \color{blue}+4.59	&  \color{blue}+4.19\\ 
    		right foot 			& \color{blue}+0.08 	& \color{blue}+4.62	\\ \hline
    		left thigh		& \color{red}-4.74 &  \color{blue}+4.15 \\ 
    		left leg 			 & \color{red}-3.80 & \color{blue}+4.54	\\ 
    		left foot 			& \color{red}-0.36	& \color{blue}+2.08 	\\ \hline
    \end{tabular}
\end{table}

\section{EXPERIMENT}

To demonstrate the performance of 
the adversarial body shape search
in \textbf{Algorithm} \ref{alg1},
we conducted experiments on a physical simulation environment MuJoCo \cite{todorov2012mujoco} with 
three legged robots in OpenAI Gym \cite{brockman2016openai}: Walker2d-v2, Ant-v2, and Humanoid-v2, as shown in Fig.~\ref{robot}.

\begin{figure*}[t]
  \begin{center}
    \begin{tabular}{c}
    	\begin{minipage}[h]{0.185\linewidth}
    		\centering
    		\includegraphics[width=0.98\columnwidth]{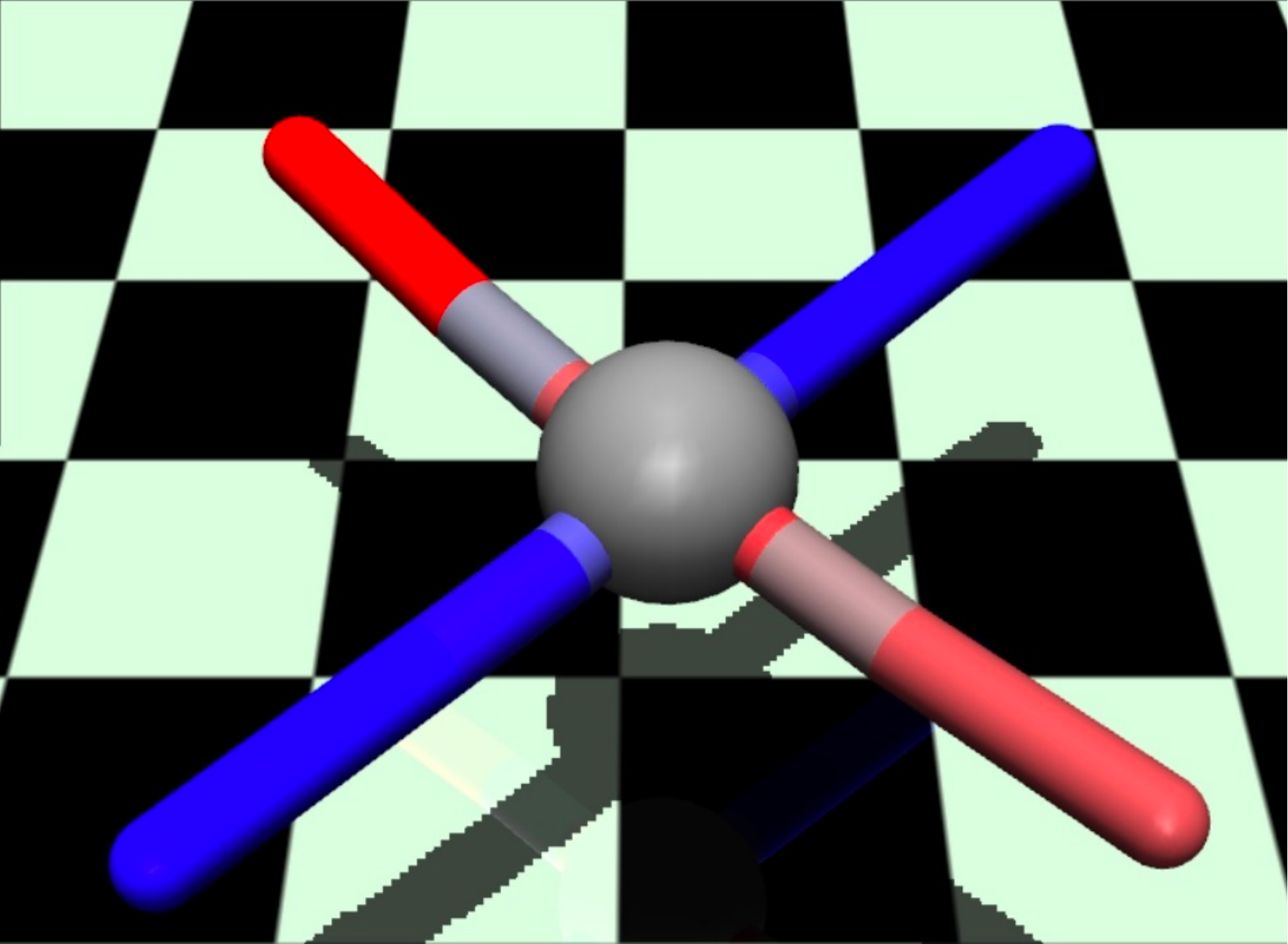}
        \subcaption{length}
    		\includegraphics[width=0.98\columnwidth]{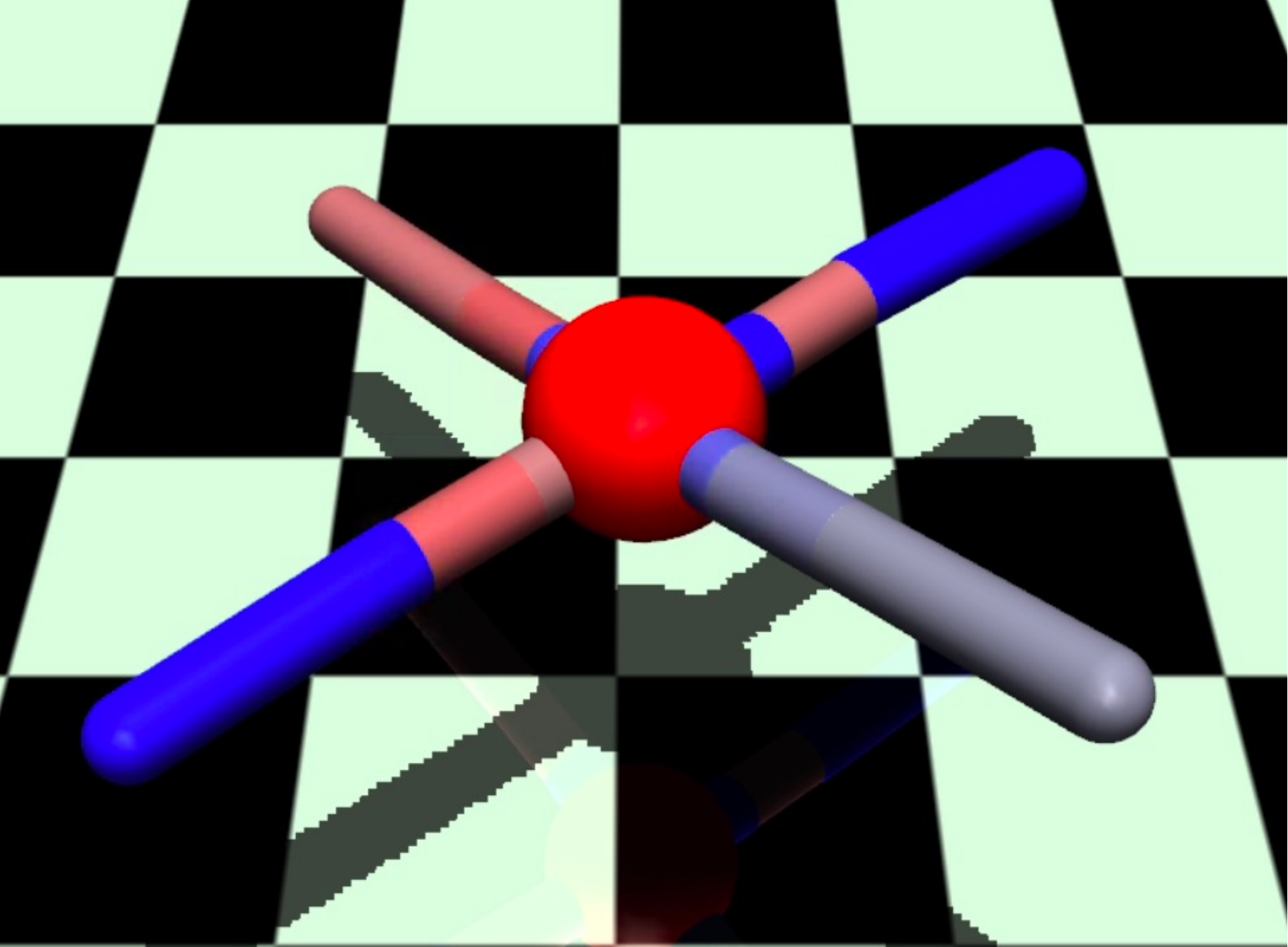}
        \subcaption{thickness}
    	\end{minipage}%
    	\begin{minipage}[h]{0.805\linewidth}
    		\centering
    	    \includegraphics[width=0.98\columnwidth]{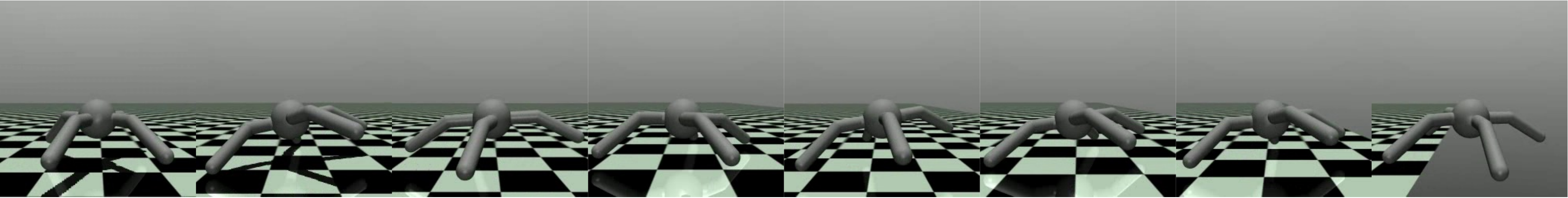}
            \includegraphics[width=0.98\columnwidth]{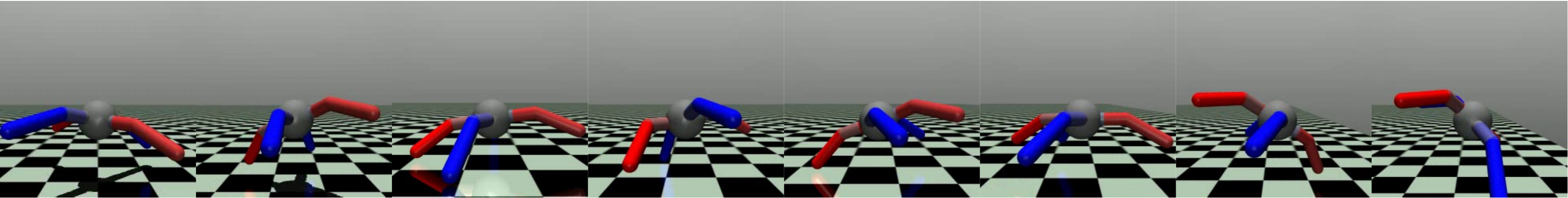}
            \includegraphics[width=0.98\columnwidth]{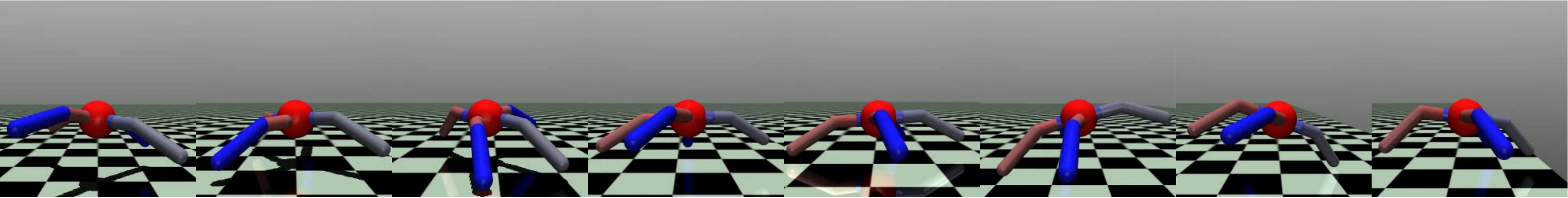}
       \subcaption{Walking animations: clean  (top), length attack  (middle), thickness attack  (bottom)}
       \label{multi pic ant}
       \end{minipage}	
    \end{tabular}
    \caption{Adversarial body shapes with length and thickness perturbations for Ant-v2. }
    \label{ant}
  \end{center}
\end{figure*}

\begin{figure}[t]
    \begin{center}
        \includegraphics[width=0.9\linewidth]{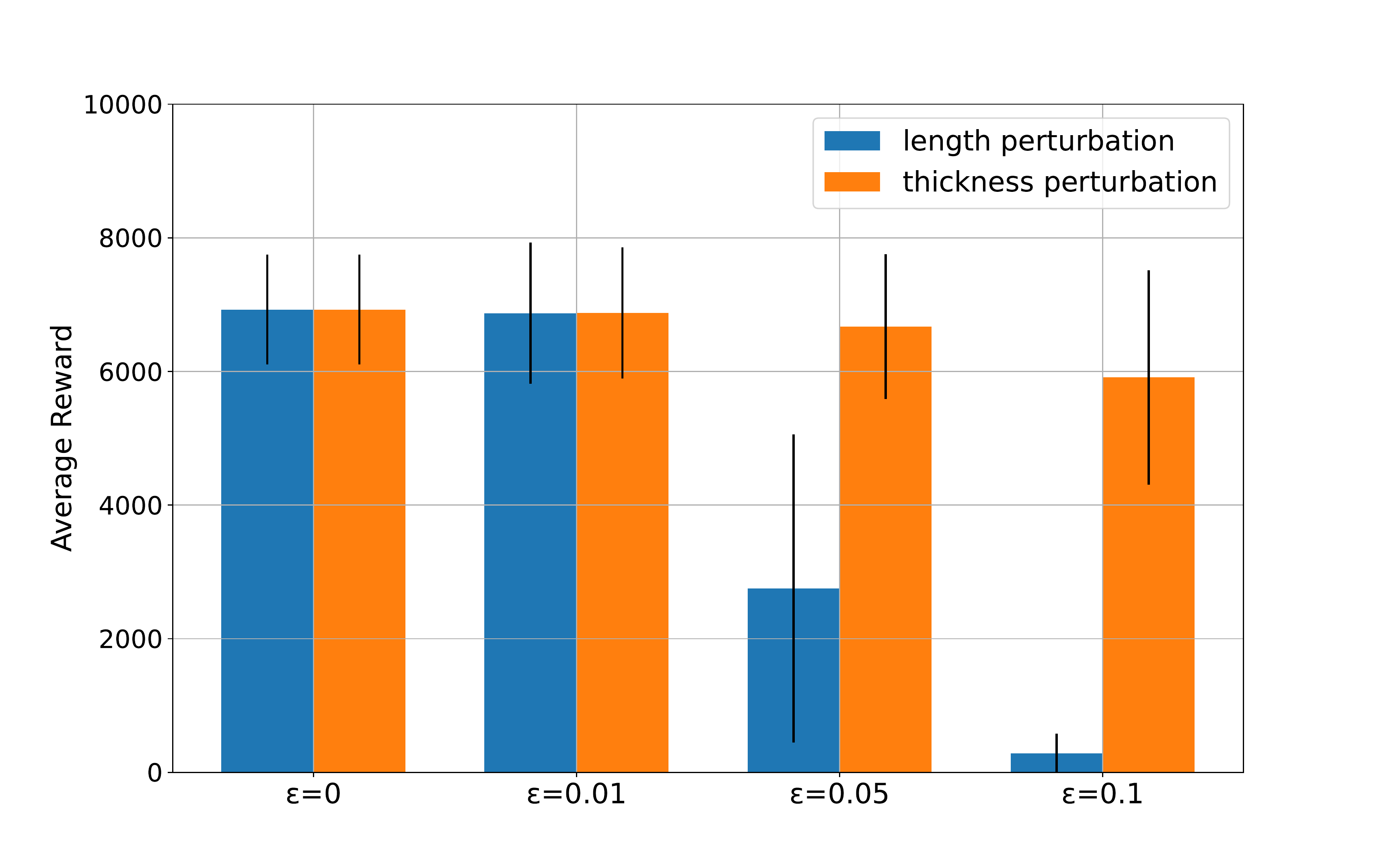}
    	\caption{Averages of the average cumulative rewards
    	with attack strength $\epsilon=0.0,0.0.1,0.05,0.1$
    	for Ant-2d}
    	\label{bar ant}
    \end{center}
    
\end{figure}

\subsection{Experimental setup}
Before the experiments,
we first trained these three robots until 
they can walk well
using the proximal policy optimization (PPO) \cite{schulman2017proximal}.
The PPO is a policy gradient-based reinforcement
learning algorithm and suitable for continuous control.
We set hyperparameters for training as follows:
learning rate is $3.0\cdot10^{-5}$, epochs are $10$, clip range is
$0.1$, batch size is $256$, horizon is $2048$. 
We designed reward functions to ensure that the robots do not fall and walk along the $x$-axis as fast as possible.
The respective reward functions for Walker2d-v2, Ant-v2,
and Humanoid-v2 are given as
follows:
\begin{align}
	r_t &= v_{fwd} - 10^{-3}||\bm{u}||^2 + 1.0,\\
    r_t &= v_{fwd} - 0.5||\bm{u}||^2 - 0.5\cdot10^{-3}||\bm{f}||^2 + 1.0,\\
    r_t &= 5.0\ v_{fwd} - 0.1||\bm{u}||^2 - 0.5\cdot10^{-6}||\bm{f}||^2 + 4.0,
\end{align}
where $v_{fwd}$ is the forward velocity, $\bm{u}$ is the vector of joint torques, and $\bm{f}$ is the impact force vector.

For performance evaluation,
we compute average cumulative rewards $\bar{C}(\bm{b}_\mathrm{adv})$
in Eq.~(\ref{equ: average cumulative reward}) 1000 times and then average them as 
\begin{equation}
    \frac{1}{1000}\sum_{i=1}^{1000}\bar{C}^{(i)}(\bm{b}_\mathrm{adv}),
    \label{equ: average average reward}
\end{equation}
where $\bar{C}^{(i)}(\bm{b}_\mathrm{adv})$ is the average cumulative reward
at the $i$-th simulation run.
Note that $\bar{C}(\bm{b}_\mathrm{adv})$ can vary with each simulation run
even if $\bm{b}_\mathrm{adv}$ is fixed,
because action $\bm{a}_t$ is stochastically generated according to policy
network $\pi(\bm{a}_t|\bm{s}_t)$.
To compute $\bar{C}(\bm{b}_\mathrm{adv})$,
we set the attack strength as $\epsilon=0.0, 0.01, 0.05, 0.1$,
where $\epsilon=0.0$ means no attack.
Following the literature on adversarial attack, 
we call the body shape with $\epsilon=0.0$ 
the \textit{clean} body, and use it as the baseline in this study.
In differential evolution, we set the number of individuals
to $NP=14, 26, 38$ for Walker2d-v2, Ant-v2, and Humanoid-v2, respectively,
and set the number of the generations to $100$ for all the robots.

\subsection{Adversarial body shape search}
We investigated the performance evaluation for each legged
robots: Walker2d-v2, Ant-v2, and Humanoid-v2.

\subsubsection{Walker2d-v2}\label{W}

Figure \ref{walker2d} visualizes the perturbations
of the length and thickness with $\epsilon= 0.05$.
The blue body parts represent those that are longer or thicker
than the clean ones, i.e., positive perturbations
are added, 
while the red ones are those that are shorter or thinner
with negative perturbations.
The darker the color becomes, the greater the perturbation
value becomes.
Table \ref{ratio walker2d} shows the perturbation values
for each body part of Fig.~\ref{walker2d}.
From the table, we find that the right and left
sides of the whole body are longer and shorter, respectively.
This result suggests that the attack on the length
breaks the left-right symmetry of the body shape
and throws the robot off balance.
On the other hand, the table indicates that
all  the thicknesses increase, i.e.,
the attack on the thickness makes
the robot heavier and more difficult to move than the clean one.

\begin{table}[t]
    \caption{Adversarial body shape perturbation for thickness
and length of Ant-v2}
    \label{ratio ant}
    \centering
    \begin{tabular}{|l||r|r|c|} \hline
    	body part	& length (\%)	& thickness(\%)	\\ \hline \hline
    	torso 			 & - \ \ \  & \color{red}-4.51		\\ \hline
    	left front thigh 			 & \color{red}-1.88 & \color{blue}+2.95\\ 
    	left front leg  & \color{blue}+0.30 & \color{red}-1.83 \\ 
    	left front foot 			& \color{red}-4.68 	& \color{red}-1.28 	\\ \hline
    	right front thigh		 & \color{blue}+3.13 & \color{blue}+4.29 \\ 
    	right front leg			 & \color{blue}+4.70& \color{red}-1.41 	\\ 
    	right front foot			 & \color{blue}+4.85 & \color{blue}+4.38		\\ \hline
    	left back thigh			 & \color{blue}+2.05 & \color{red}-1.06\\ 
    	left back leg			& \color{blue}+4.17 & \color{red}-1.76  	\\ 
    	left back foot			& \color{blue}+4.90 	& \color{blue}+4.12 	\\ \hline
    	right back thigh			& \color{red}-2.67 & \color{blue}+1.94 \\ 
    	right back leg			 & \color{red}-0.54 & \color{blue}+0.67 	\\ 
    	right back foot			& \color{red}-2.16 & \color{blue}+0.36 		\\ \hline
    \end{tabular}
\end{table}

\begin{figure*}[t]
  \begin{center}
    \begin{tabular}{c}
    	\begin{minipage}[h]{0.108\linewidth}
    		\centering
    		\includegraphics[width=0.98\columnwidth]{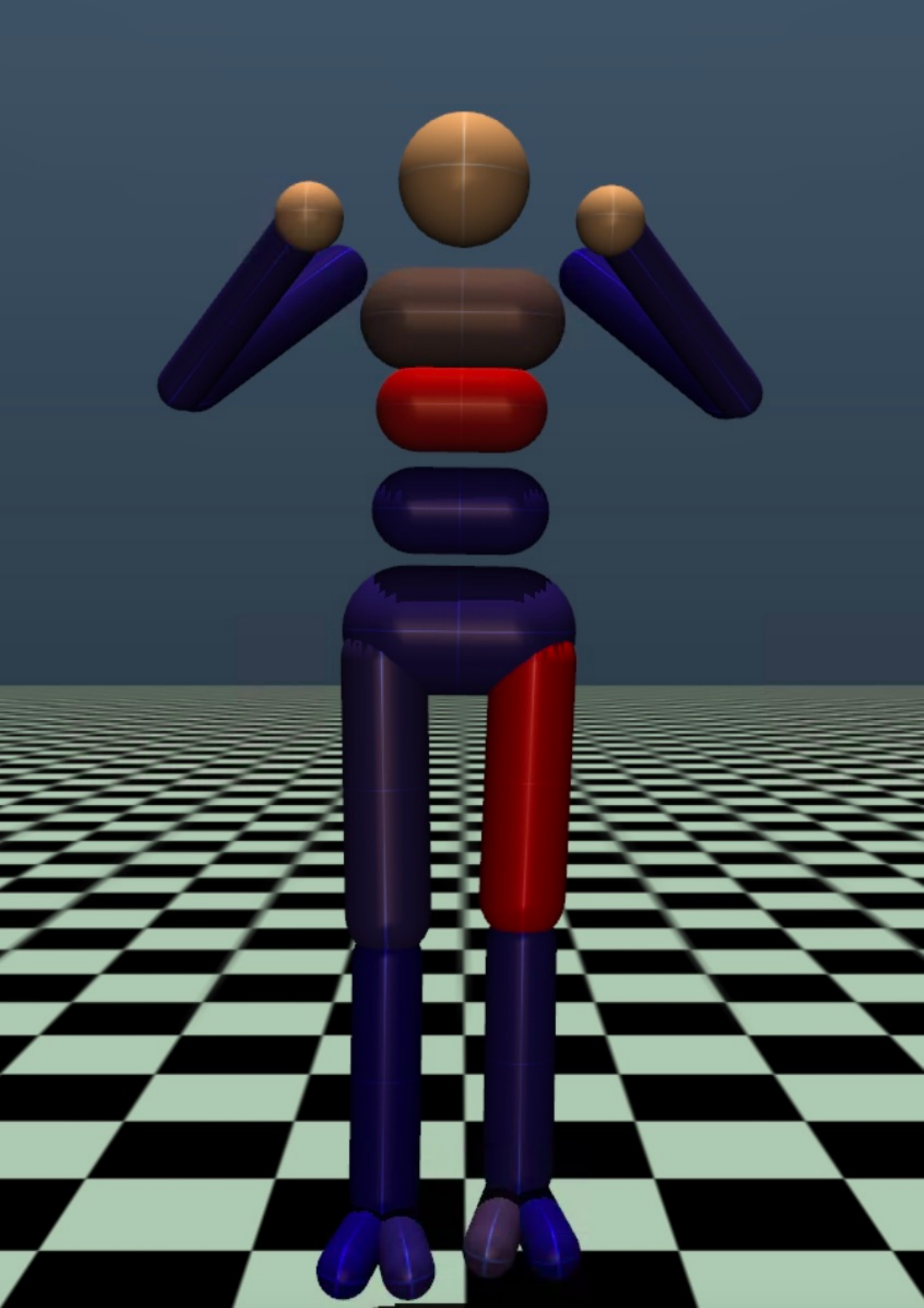}
        \subcaption{length}
    		\includegraphics[width=0.98\columnwidth]{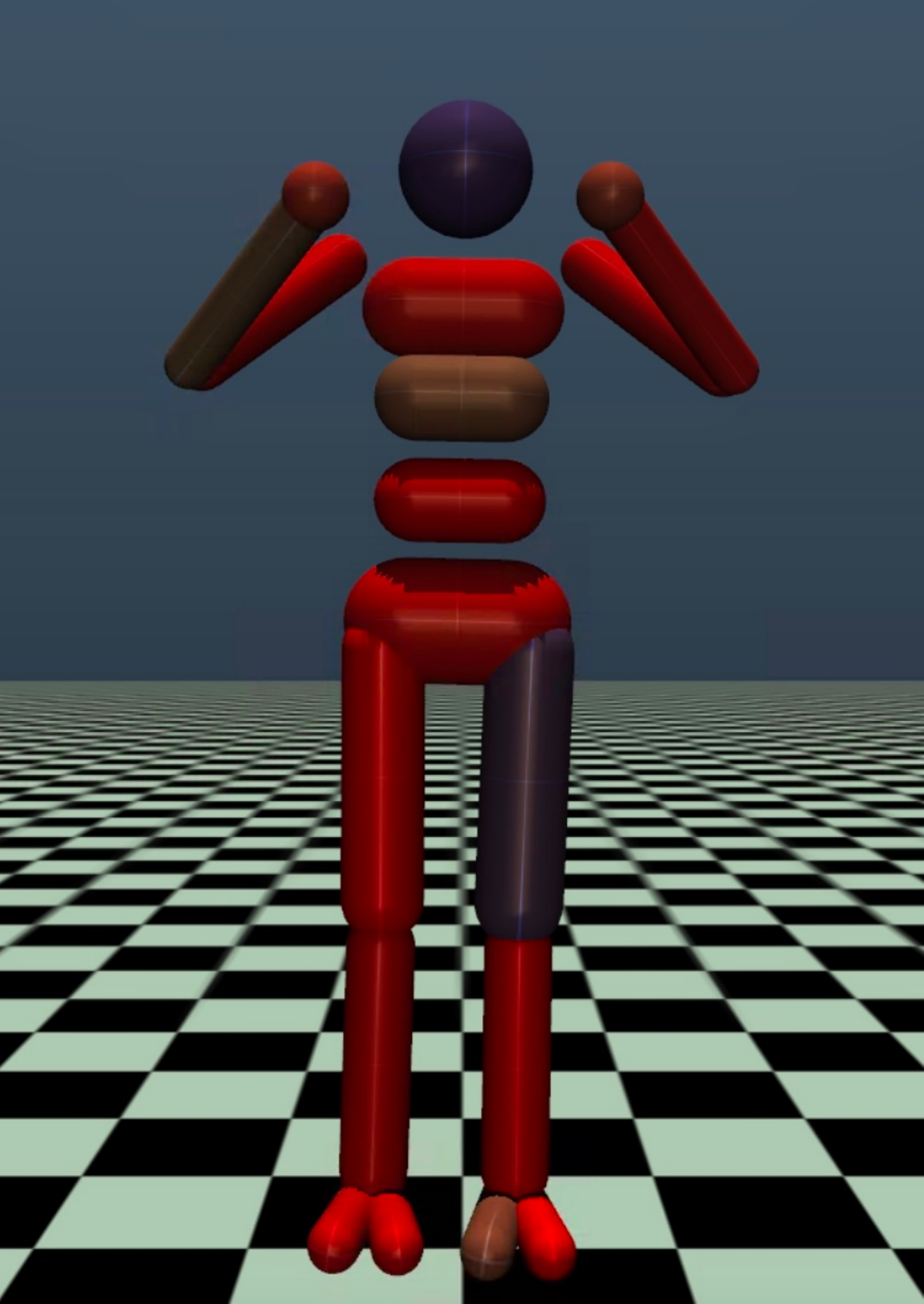}
        \subcaption{thickness}
    	\end{minipage}%
    	\begin{minipage}[h]{0.892\linewidth}
    		\centering
    	    \includegraphics[width=0.98\columnwidth]{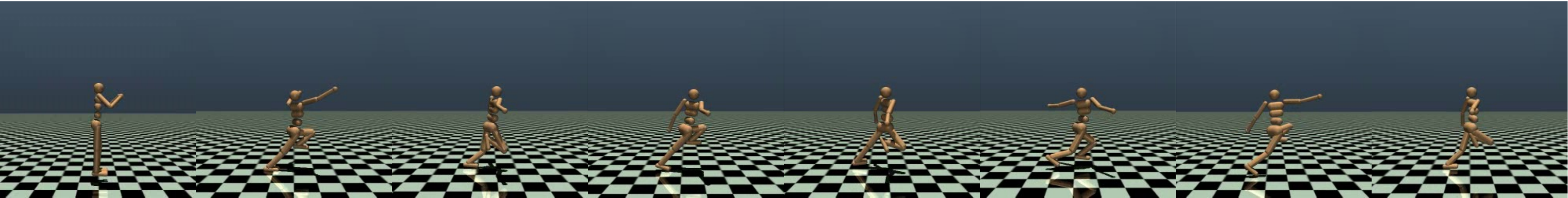}
            \includegraphics[width=0.98\columnwidth]{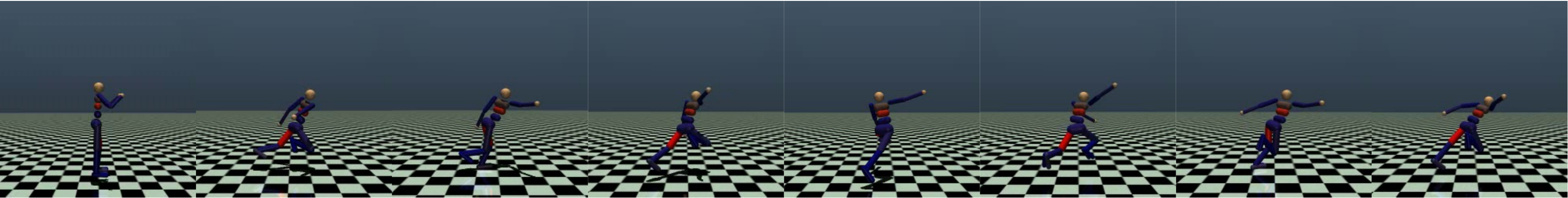}
            \includegraphics[width=0.98\columnwidth]{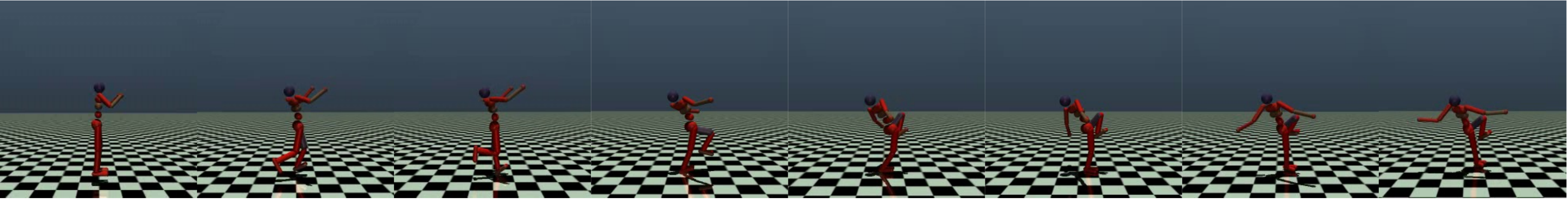}
       \subcaption{Walking animations: clean  (top), length attack  (middle), thickness attack  (bottom)}
       \label{multi pic humanoid}
       \end{minipage}	
    \end{tabular}
    \caption{Adversarial body shapes with length and thickness perturbations for Humanoid-v2. }
    \label{humanoid}
  \end{center}
\end{figure*}

\begin{figure}[t]
    \centering
    \includegraphics[width=0.9\linewidth]{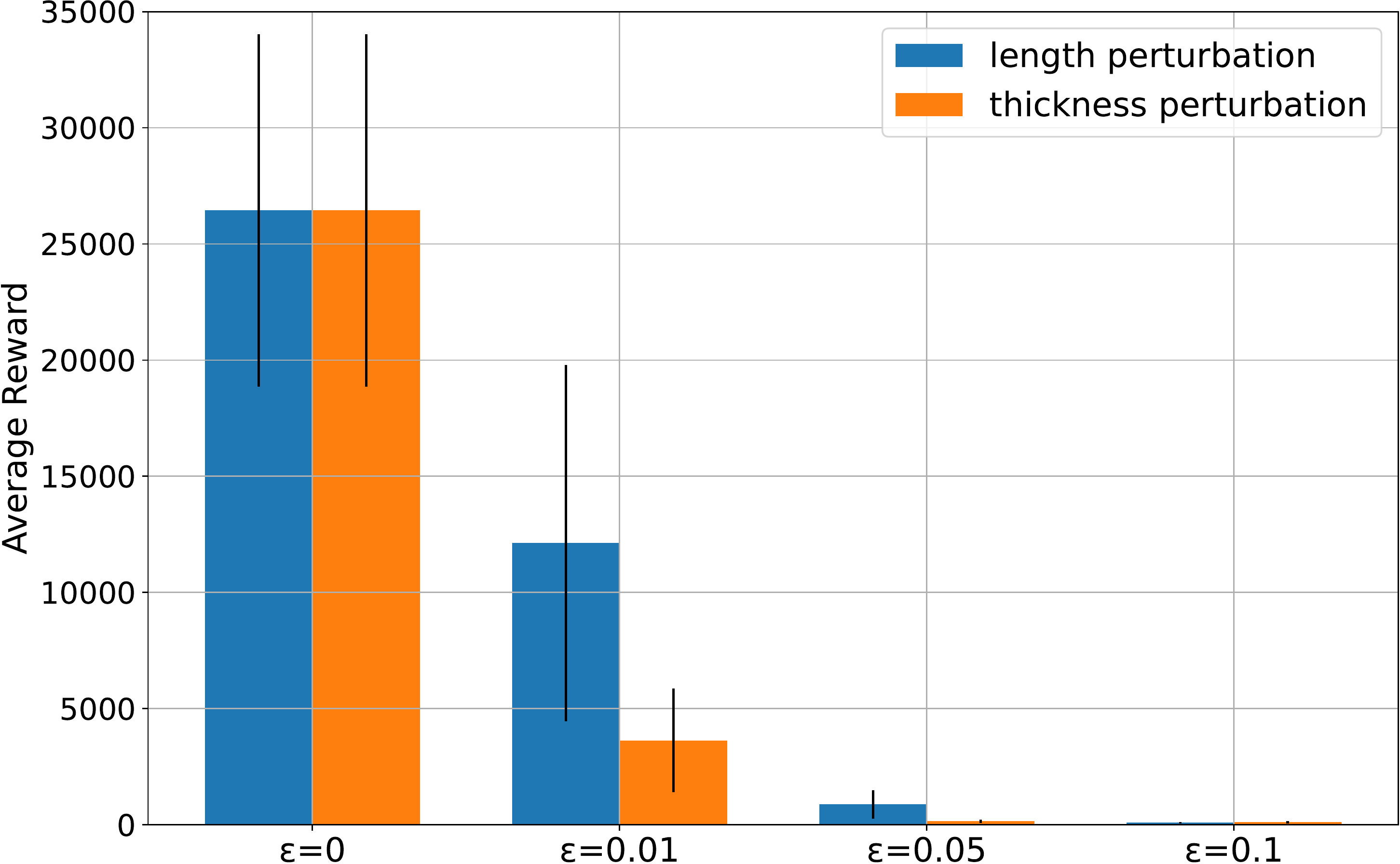}
    \caption{Averages of the average cumulative rewards
    	with attack strength $\epsilon=0.0,0.0.1,0.05,0.1$
    	for Humanoid-2d}
    \label{bar humanoid}
\end{figure}

Figure \ref{multi pic walker2d} shows the animations of
the walking simulations for clean (top),
length attack (middle), and thickness attack (bottom).
We see that the length-attacked robot (middle) loses 
its balance and falls.
This result comes from the 
left-right asymmetry of the body, as shown in
Fig.~\ref{walker2d} and Table \ref{ratio walker2d}.
We also see that the thickness-attacked robot (bottom) 
falls. 
As mentioned above, 
the thickness attack weighs down the robot; thus,
it becomes difficult for the robot to move the legs for stable walking.

Figure \ref{bar walker2d} shows the \textit{average} of the 1000 average 
cumulative rewards in Eq.~(\ref{equ: average average reward})
for Walker2d-v2.
In Fig.~\ref{bar walker2d}, the blue and orange bars show
the average rewards of the attacks on
the length and thickness, respectively.
The results demonstrate that the average rewards decrease
by attacking the body shape with strength 
$\epsilon\geq 0.05$,
e.g., the length perturbation with $\epsilon= 0.05$ 
reduces the reward by 82\% to the baseline ($\epsilon=0.0$).
In particular, we find that Walker2d-v2
is more vulnerable to the attack on the length than
the thickness of the body parts.

\subsubsection{Ant-v2}\label{A}

Figure \ref{ant} visualizes the perturbations of the length and thickness with $\epsilon=0.05$. 
Like Walker2d-v2 in Fig.~\ref{walker2d},
blue and red mean the positive and negative 
perturbations, respectively.
Table \ref{ratio ant} shows the perturbation values for each body part of Fig.~\ref{ant}.
From the table, 
we find that the right front and left back parts are
longer and the other two parts are shorter than
the clean one.
As shown in Fig.~\ref{multi pic ant} (top),
the clean Ant-v2 uses its right front and left back legs
to move forward and uses the other legs to support its body.
Hence, the adversarial perturbations in Table \ref{ratio ant} 
cause a pitch oscillation of Ant-v2, as shown in
Fig.~\ref{multi pic ant} (middle).
Unlike the length perturbation, 
the thickness perturbation has no significant effect on
walking, as shown in Fig.~\ref{multi pic ant} (bottom).

\begin{table}[t]
    \caption{Adversarial body shape perturbation for thickness
and length of Humanoid-v2}
    \label{ratio humanoid}
    \centering
    \begin{tabular}{|l||r|r|r|} \hline
    	body part		& length (\%) & thickness (\%)	\\ \hline \hline
        head 			  & - \ \ \	 & \color{blue}+2.21   \\ \hline
        torso 			 & \color{blue}+0.69 &  \color{red}-3.11 		\\ 
    	upper waist 			 & \color{blue}+3.71 	& \color{red}-0.39 	\\ 
    	lower waist 			 & \color{red}-2.89 	& \color{red}-3.78 	\\ 
    	pelvis 			  &  \color{blue}+3.36 & \color{red}-3.38		\\ \hline
    	right thigh 			  &  \color{blue}+2.96 &   \color{red}-4.91		\\ 
        right leg 			  &  \color{blue}+4.48 	&  \color{red}-4.09	\\ 
    	right right foot 			 &  \color{blue}+4.65 &  \color{red}-3.43 		\\ 
    	right left foot 			 &  \color{blue}+3.46 &  \color{red}-3.85 		\\ \hline
    	left thigh 			 & \color{red}-4.54 	&  \color{blue}+1.66 	\\ 
    	left leg 			 &   \color{blue}+3.84 	&  \color{red}-4.91 	\\ 
        left right foot 			 &  \color{blue}+2.16	&  \color{red}-0.93 	\\ 
    	left left foot 			 &  \color{blue}+4.41 &  \color{red}-4.86 		\\ \hline
    	right upper arm 			  &  \color{blue}+3.69 &  \color{red}-3.29		\\ 
    	right lower arm 			  &  \color{blue}+3.35 	&  \color{blue}+0.05	\\ 
    	right hand 			  & -  \ \ \		& \color{red}-1.86	\\ \hline
        left upper arm  			  &  \color{blue}+4.09	&  \color{red}-3.09	\\ 
    	left lower arm 			 &  \color{blue}+3.24 &  \color{red}-2.9 		\\ 
    	left hand 			  & - 	 \ \ \	&  \color{red}-1.27	\\ \hline
    \end{tabular}
\end{table}

Figure \ref{bar ant} shows the average of the 1000
average cumulative rewards in Eq.(\ref{equ: average average reward}) for Ant-v2. 
The results demonstrate that the length attack is
successful with $\epsilon\geq0.05$, whereas 
the thickness attack is not.
For example, the length attack with $\epsilon=0.05$
reduces the average reward by 60\% to that of the clean one,
and the thickness attack does not.
These results indicate that Ant-v2 is vulnerable to 
the length attack and is robust to the thickness attack.

\subsubsection{Humanoid-v2}\label{H}

Figure \ref{humanoid} visualizes the perturbations of the length and
thickness with $\epsilon=0.05$.
Table \ref{ratio humanoid} shows the perturbation values for each body part of Fig.~\ref{humanoid}.
From the data in the table, we infer that the right leg parts---thigh, leg, right foot, and left foot---
are longer and the left thigh is shorter than those of the clean one by attacking the lengths.
Like that observed for Walker2d-v2,
these adversarial length perturbations break the left-right symmetry of the body shape and throws
the robot off balance, as shown in Fig.~\ref{multi pic humanoid} (middle).
On the other hand, Table \ref{ratio humanoid} lists the thickness perturbations that make
the head larger and the other parts thinner.
As shown in Fig.~\ref{multi pic humanoid} (bottom),
the head of Humanoid-v2 is so heavy that it loses its balance and is pulled backward.

Figure \ref{bar humanoid} shows the average of the 1000 average cumulative rewards
in Eq.~(\ref{equ: average average reward}) for Humanoid-v2. 
The results demonstrate that 
Humanoid-v2 is vulnerable to the adversarial attack on both the length and thickness, 
in relation to Walker2d-v2 and Ant-v2.
Hence, both attacks can significantly reduce the average rewards with even $\epsilon\geq0.01$,
and the average rewards eventually become almost zero when $\epsilon=0.1$.

\section{CONCLUSION}
We proposed an evolutionary computation method for searching adversarial body
shapes of the legged robots. 
The vulnerability to small body changes can be a potentially significant risk,
because they cause the robots to fall.
Because deep reinforcement learning has been widely used in robotics,
finding vulnerability is very important for the
safety and robustness in robotics.
This study demonstrated that the legged robots—Walker2d, Ant-v2, and Humanoid-v2—
are vulnerable to the attacks on the body shape and can be forced to fall.
Through the experiments, we revealed that the adversarial attacks perturb the length or 
thickness of the body parts such that the left-right symmetry is broken or 
the center of gravity is shifted.
In future, we will develop a method to design robust body shapes 
against the adversarial attacks to improve the safety and robustness of legged robots.


\bibliographystyle{unsrt}
\bibliography{refs}

\begin{thebibliography}{10}

\bibitem{10.1007/s11370-021-00398-z}
Eduardo~F. Morales, Rafael Murrieta-Cid, Israel Becerra, and Marco~A.
  Esquivel-Basaldua.
\newblock A survey on deep learning and deep reinforcement learning in robotics
  with a tutorial on deep reinforcement learning.
\newblock {\em Intell. Serv. Robot.}, 14(5):773^^e2^^80^^93805, nov 2021.

\bibitem{xiao2019characterizing}
Chaowei Xiao, Xinlei Pan, Warren He, Jian Peng, Mingjie Sun, Jinfeng Yi,
  Mingyan Liu, Bo~Li, and Dawn Song.
\newblock Characterizing attacks on deep reinforcement learning.
\newblock {\em arXiv preprint arXiv:1907.09470}, 2019.

\bibitem{9536399}
Inaam Ilahi, Muhammad Usama, Junaid Qadir, Muhammad~Umar Janjua, Ala Al-Fuqaha,
  Dinh~Thai Hoang, and Dusit Niyato.
\newblock Challenges and countermeasures for adversarial attacks on deep
  reinforcement learning.
\newblock {\em IEEE Transactions on Artificial Intelligence}, 3(2):90--109,
  2022.

\bibitem{storn1997differential}
Rainer Storn and Kenneth Price.
\newblock Differential evolution--a simple and efficient heuristic for global
  optimization over continuous spaces.
\newblock {\em Journal of global optimization}, 11(4):341--359, 1997.

\bibitem{brockman2016openai}
Greg Brockman, Vicki Cheung, Ludwig Pettersson, Jonas Schneider, John Schulman,
  Jie Tang, and Wojciech Zaremba.
\newblock Openai gym, 2016.

\bibitem{todorov2012mujoco}
Emanuel Todorov, Tom Erez, and Yuval Tassa.
\newblock Mujoco: A physics engine for model-based control.
\newblock In {\em 2012 IEEE/RSJ International Conference on Intelligent Robots
  and Systems}, pages 5026--5033. IEEE, 2012.

\bibitem{Ha2019ReinforcementLF}
David~R Ha.
\newblock Reinforcement learning for improving agent design.
\newblock {\em Artificial Life}, 25:352--365, 2019.

\bibitem{Schaff2019JointlyLT}
Charles~B. Schaff, David Yunis, Ayan Chakrabarti, and Matthew~R. Walter.
\newblock Jointly learning to construct and control agents using deep
  reinforcement learning.
\newblock {\em 2019 International Conference on Robotics and Automation
  (ICRA)}, pages 9798--9805, 2019.

\bibitem{pmlr-v100-luck20a}
Kevin~Sebastian Luck, Heni~Ben Amor, and Roberto Calandra.
\newblock Data-efficient co-adaptation of morphology and behaviour with deep
  reinforcement learning.
\newblock In {\em Proceedings of the Conference on Robot Learning}, pages
  854--869, 2020.

\bibitem{wang2018neural}
Tingwu Wang, Yuhao Zhou, Sanja Fidler, and Jimmy Ba.
\newblock Neural graph evolution: Automatic robot design.
\newblock In {\em International Conference on Learning Representations}, 2019.

\bibitem{wang2018nervenet}
Tingwu Wang, Renjie Liao, Jimmy Ba, and Sanja Fidler.
\newblock Nervenet: Learning structured policy with graph neural networks.
\newblock In {\em International Conference on Learning Representations}, 2018.

\bibitem{DBLP:journals/corr/abs-1801-00385}
Ruta Desai, Beichen Li, Ye~Yuan, and Stelian Coros.
\newblock Interactive co-design of form and function for legged robots using
  the adjoint method.
\newblock {\em CoRR}, abs/1801.00385, 2018.

\bibitem{schulman2017proximal}
John Schulman, Filip Wolski, Prafulla Dhariwal, Alec Radford, and Oleg Klimov.
\newblock Proximal policy optimization algorithms.
\newblock {\em arXiv preprint arXiv:1707.06347}, 2017.

\end{thebibliography}

\end{document}